\documentclass[sigconf,nonacm]{acmart}

\renewcommand\footnotetextcopyrightpermission[1]{}

\AtBeginDocument{%
  }

\usepackage{colortbl}
\definecolor{bgTrajMind}{HTML}{E8F3FC}
\definecolor{bgTableSection}{HTML}{F2F2F2}
\usepackage{placeins}
\usepackage{algorithm}
\usepackage{algpseudocode}
\usepackage{tabularx}
\usepackage{longtable}
\usepackage{pdflscape}
\newcolumntype{Y}{>{\centering\arraybackslash}X}

\AtBeginDocument{%
  \setlength{\abovedisplayskip}{4pt plus 1.5pt minus 1.5pt}%
  \setlength{\belowdisplayskip}{4pt plus 1.5pt minus 1.5pt}%
  \setlength{\abovedisplayshortskip}{2pt plus 2pt minus 1pt}%
  \setlength{\belowdisplayshortskip}{3pt plus 2pt minus 1pt}%
  \setlength{\parskip}{0pt}%
}
\usepackage{pgfplots}
\pgfplotsset{compat=1.18}
\usetikzlibrary{patterns,arrows.meta,positioning}

\graphicspath{{figures/}}

\begin{document}

\flushbottom

\title[TrajMind]{TrajMind: Chaining Role-Specialized LoRAs for Fast-and-Slow
Collective Trajectory Anomaly Diagnosis}

\author{Jiahao Wu}
\email{jiahao.wu@connect.polyu.hk}
\affiliation{%
  \institution{The Hong Kong Polytechnic University}
  \city{Hong Kong}
  \country{China}}

\author{Zhen-qun Yang}
\email{look.jessica@gmail.com}
\affiliation{%
  \institution{The Hong Kong Polytechnic University}
  \city{Hong Kong}
  \country{China}}

\author{Chen Jason Zhang}
\email{jason-c.zhang@polyu.edu.hk}
\affiliation{%
  \institution{The Hong Kong Polytechnic University}
  \city{Hong Kong}
  \country{China}}

\author{Qing Li}
\email{csqli@comp.polyu.edu.hk}
\affiliation{%
  \institution{The Hong Kong Polytechnic University}
  \city{Hong Kong}
  \country{China}}

\renewcommand{\shortauthors}{Wu et al.}

\begin{abstract}
Diagnosing collective anomalies from urban trajectories is increasingly important for traffic governance, as it reveals what happened, who was involved, and where and when the event occurred. 
Existing detectors efficiently produce scores or labels, whereas vision--language pipelines provide richer semantics; neither couples verifiable diagnosis with low-latency monitoring. 
The central challenge is to recognize collective patterns and recover exact
event details from the source trajectories without running the full diagnostic
pipeline for every monitored window. 
We therefore separate always-on screening from on-demand diagnosis: screening raises alerts, while diagnosis
releases only source-verified what--who--where--when records. 
We present 
TrajMind, a fast-and-slow framework that switches three role-specialized LoRA
adapters over one frozen vision--language backbone. 
Its slow path, \textit{TrajMind$_{\text{slow}}$}, chains canvas-based typing, 
type-conditioned localization over serialized trajectories, and executable
verification, yielding structured, evidence-backed diagnoses. 
Additionally, the fast path, \textit{TrajMind$_{\text{fast}}$}, screens each window in a single text-only pass, delivering efficient structured alerts. Extensive experiments show that, TrajMind$_{\mathrm{slow}}$ outperforms the strongest baselines by at least $15.3$ percentage points in anomaly typing and $13.8$ percentage points in localization. These gains persist under cross-city transfer, and TrajMind$_{\mathrm{fast}}$ reduces latency by $41.1\%$ and maintains binary balanced accuracy of at least $93.5\%$. 
Together, TrajMind delivers accurate, evidence-backed diagnoses
across cities and efficient front-line monitoring.
\end{abstract}


\settopmatter{printccs=false}

\maketitle

\section{Introduction}

Collective trajectory anomaly detection and diagnosis are becoming indispensable to modern urban traffic governance. 
Developing effective anomaly detection and diagnosis methods is crucial for ensuring traffic safety, enhancing the overall efficiency of urban transportation systems and has become an increasingly important research topic~\citep{zhang2011ibat,han2022deeptea,bu2025datmamba}. However, turning the trajectories into operational evidence requires more than assigning an anomaly score to an isolated trip, which is far from what existing methods can accomplish~\citep{zhang2011ibat,liu2020online,han2022deeptea,wang2024mstoatd}. 
Given a window of map-matched trajectories, a collective anomaly system must reason over agent interactions to detect abnormal collective behavior and identify its type, participants, affected road segments, and duration. Operational systems must do so accurately while keeping pace with continuously arriving windows. Building a practical model capable of accurate diagnosis and efficient monitoring remains a fundamental challenge.

Classical trajectory anomaly detectors define normality through route
similarity, density, or isolation \citep{lee2008trajectory,zhang2011ibat}.
Although efficient, these detectors rely on route frequency, distance, or reference statistics that vary across locations and traffic conditions, and therefore generalize poorly without per-site recalibration.  They also focus primarily on isolated trajectories rather than events produced by interactions
among co-present road users.  Deep sequence models capture time-dependent route distributions \citep{liu2020online,han2022deeptea}, while graph-based
approaches model social dependencies among nearby vehicles
\citep{bu2025datmamba}.  These richer representations improve the ability to
recognize complex motion patterns, but they remain optimized mainly for
producing an anomaly score or label.  Therefore, these methods fails to identify the involved event participants,
and ground the event in space and time with evidence.  

Recently, vision-language models (VLMs) have introduced semantic reasoning into trajectory anomaly analysis~\citep{liu2025trajmllm,zanella2024lavad,zhang2025holmes}. For instance, Traj-MLLM renders an individual trajectory and its map context as interleaved multiview image--text inputs for training-free multimodal reasoning~\citep{liu2025trajmllm}. However, this paradigm remains limited for operational group anomaly diagnosis. By reasoning largely over one trajectory at a time, it may miss anomalies that emerge only from the timing and co-occurrence of multiple agents, and it does not explicitly identify the anomaly type, involved agents, affected road segments, and time span. Moreover, multiview rendering and repeated invocation of a large multimodal backbone introduce substantial latency, making always-on monitoring costly~\citep{mao2026vibes}. Thus, although VLMs enable richer semantic reasoning, existing VLM pipelines are neither sufficiently group-aware nor efficient for continuous trajectory anomaly diagnosis.

To address these challenges, we propose TrajMind, a VLM framework with specialized roles for collective trajectory anomaly detection and diagnosis. Our key insight is that the task decomposes into three heterogeneous capabilities:
{scene-level understanding} to determine \emph{what} happened,
{precise grounding} to recover \emph{who}, \emph{where}, and
\emph{when}, and {efficient monitoring} to decide whether an incoming
window warrants attention. Supporting all three capabilities simultaneously
entails \emph{conflicting representational requirements}:
\begingroup
\setlength{\leftmargini}{1.2em}
\begin{itemize}
    \setlength{\itemsep}{1pt}
    \setlength{\parsep}{0pt}
    \setlength{\topsep}{1pt}
    \setlength{\partopsep}{0pt}
    \item Scene-level understanding benefits from a holistic yet lossy visual rendering of the window, which captures the overall traffic pattern and context.
    \item Precise grounding requires lossless textual access to agent identities, coordinates, and timestamps, enabling accurate identification of involved participants and their spatiotemporal details.
    \item Efficient monitoring must forgo visual processing to sustain throughput, requiring to quickly screen incoming windows.
\end{itemize}
\endgroup
{These conflicting requirements motivate role specialization rather than a single
shared objective.} TrajMind therefore assigns each capability to
a lightweight role-specific LoRA adapter over a common frozen VLM backbone
\citep{hu2022lora}.

TrajMind composes these adapters into two operating modes.  For in-depth
diagnosis, TrajMind$_{\mathrm{slow}}$ chains scene typing on a map-aligned
canvas with grounding over lossless textual records to produce a complete,
evidence-checked diagnosis.  For
monitoring, TrajMind$_{\mathrm{fast}}$ invokes only the monitoring adapter,
dropping visual processing and staged reasoning altogether and screening each
window in a single text-only pass.  Unlike
generic multi-LoRA or adapter-composition schemes that route among
interchangeable skills over a common input, TrajMind's specialists deliberately
consume different views of the same window.  This division provides detailed,
evidence-backed reasoning for in-depth investigation and an efficient
screening path for always-on monitoring. 
Our contributions are threefold:
\begingroup
\setlength{\leftmargini}{1.2em}
\begin{itemize}
    \setlength{\itemsep}{1pt}
    \setlength{\parsep}{0pt}
    \setlength{\partopsep}{0pt}
  \item \textbf{Diagnosis formulation and evaluation protocol.} We define
  structured what--who--where--when diagnoses. Our evaluation protocol distinguishes alerts from
  evidence-backed diagnoses. We evaluate this formulation on held-out
  trajectories from Chengdu, Xi'an, and Porto. Controlled collective anomalies
  provide type, participant, segment, and temporal ground truth.
  \item \textbf{Framework.} We propose TrajMind, which specializes a shared
  frozen VLM using role-specific LoRA adapters. The slow path supports anomaly
  typing and exact localization, while the fast path provides efficient
  monitoring. An executable verifier checks each diagnosis against the source
  trajectories.
  \item \textbf{Results.} Across the three cities,
  TrajMind$_{\mathrm{slow}}$ improves three-way balanced accuracy by
  $15.3$--$29.0$\% over the strongest non-TrajMind baseline.
  Under type-conditioned localization, it improves participant F1 by
  $13.8$--$23.2$\%. Results on Xi'an and Porto are zero-shot.
  TrajMind$_{\mathrm{fast}}$ reduces latency from $9.026$ to $5.315$ seconds per
  window ($41.1\%$) while retaining binary balanced accuracy of
  $.960/.949/.935$. The slow path remains stronger in fine-grained typing and participant localization.
\end{itemize}
\endgroup

\section{Related Work}


\textbf{Trajectory Anomaly Detection.} Trajectory anomaly detection identifies movement that departs from expected
spatial, temporal, or behavioral patterns~\citep{chandola2009anomaly}.  At the
\emph{individual level}, an anomaly is an unusual trajectory or motion sequence.
Classical methods detect such deviations through trajectory partitioning,
geometric similarity, density, or trajectory frequency
\citep{lee2008trajectory,zhang2011ibat}, whereas deep representation models
learn normal sequential behavior directly from data
\citep{liu2020online,han2022deeptea,jiao2023vehicle,wang2024decortad}.
At the \emph{group level}, the anomaly instead lies in relations among
co-present agents, even when each trajectory appears plausible in isolation.  Recent
methods therefore model social and temporal dependencies using recurrent graph
attention, transformers, graph Mamba, collective reconstruction, or diffusion
imputation~\citep{hu2023dsab,lohrer2024gadformer,bu2025datmamba,wen2025cobad,
ouyang2026dgad}. Self-supervised encoders and newer trajectory foundation
models further target transfer across tasks and regions
\citep{jiang2023start,zhu2025unitraj,wei2025transfertraj}. Beyond mobility,
disentangled contrastive learning separates collaborative and social factors
to transfer knowledge between heterogeneous relational views
\citep{wu2022dcrec}. This result supports preserving role-specific information,
but it does not address event-level spatiotemporal grounding. The latest
language-model approaches encode trajectories as
tokens or multimodal inputs~\citep{mbuya2024lmtad,liu2025trajmllm}, but still
center on individual trajectories or task-specific predictions.  Existing
methods consequently fail to jointly recover a collective anomaly's type,
complete participant set, affected road segments, and temporal extent.

\textbf{Vision--Language Models.}
 General-purpose vision--language models couple visual perception with language
generation and reasoning, providing adaptable backbones such as
Qwen2.5-VL~\citep{bai2025qwen25vl}.  Their scope has expanded to map and traffic
understanding~\citep{cao2024maplm}, single-trajectory mining through interleaved
map and text representations~\citep{liu2025trajmllm}, and general video dialogue
and temporal grounding~\citep{maaz2024videochatgpt,ren2024timechat}. Video anomaly
understanding has likewise advanced across temporal scales
\citep{zanella2024lavad,zhang2025holmes,mao2026vibes}. In embodied settings,
VLM adaptation has also produced generalizable rewards from learned failure
prompts and improved long-horizon decisions through value-guided multi-path
reflection~\citep{yang2024adapt2reward,yang2026seeingfarther}. In parallel,
parameter-efficient adaptation has enabled economical domain specialization
and modular reasoning through low-rank updates to frozen models
\citep{hu2022lora,huang2024lorahub,liu2026videomind}. Complementary work reduces
data cost through policy-gradient-based or LLM-driven dataset condensation
\citep{wu2025dconrec,wu2025tfdcon}, while adaptive action chunking reduces the
number of model decisions in long-horizon agents~\citep{yang2026space}.
However, extending these advances to collective trajectory diagnosis remains
an open challenge. Existing systems focus on either a single trajectory,
surveillance video, or general-purpose agent efficiency. Traffic response
instead requires a complete diagnosis of event type, participants, and
spatiotemporal extent, with outputs that remain checkable against the source
trajectories. Providing such diagnosis at the cost required for continuous
monitoring remains largely unexplored.

\section{Methodology}

\begin{figure*}[t]
\centering
\includegraphics[width=0.83\textwidth]{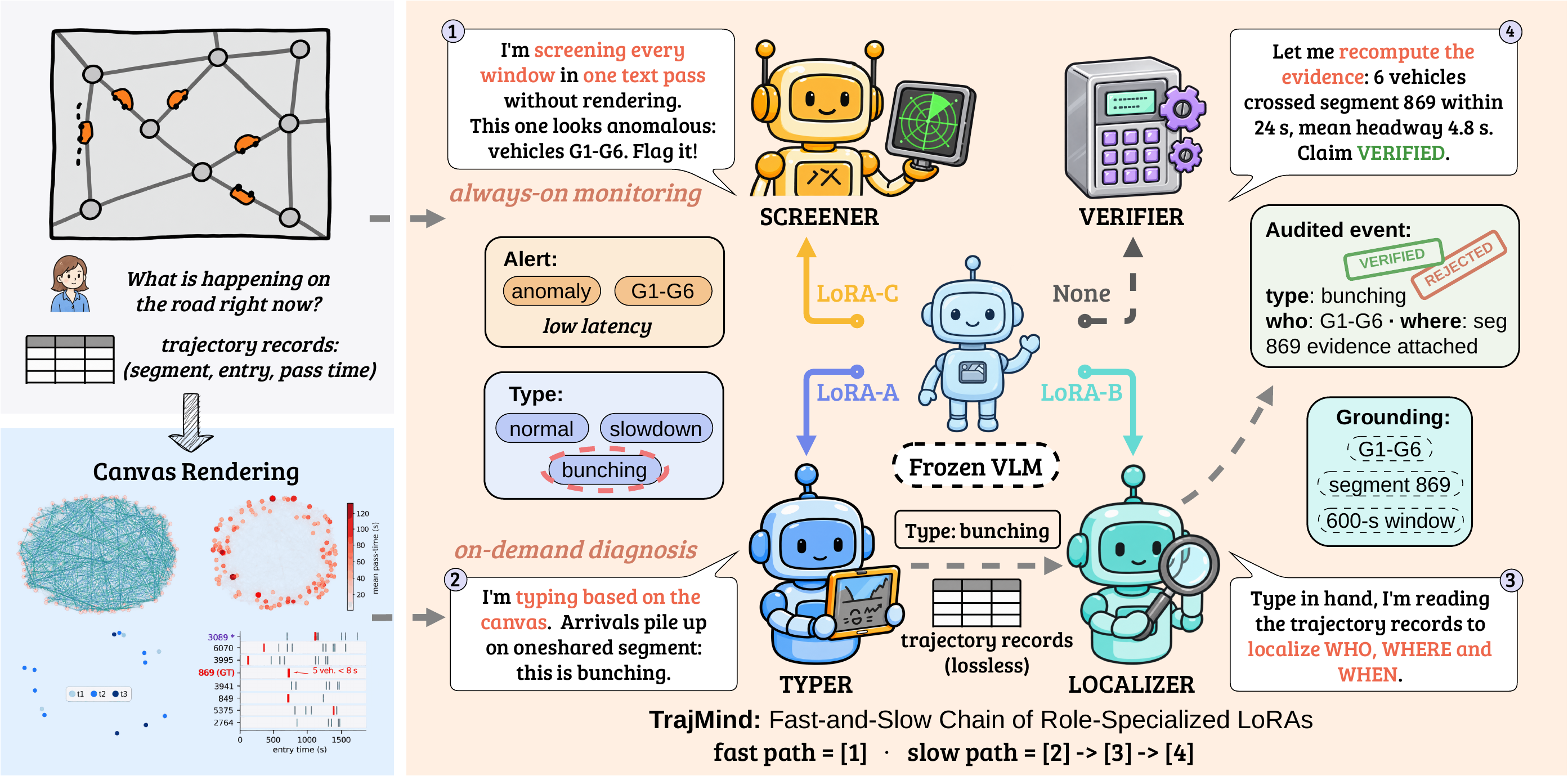}
\caption{Overview of TrajMind.  A single frozen vision--language backbone is
specialized by switching among role-specific LoRA adapters.  The
\emph{fast path} is the screener alone~(1): one text-only pass over serialized
trajectory records, with no rendering, which flags a window and names its
participants at low latency.  The \emph{slow path} chains three roles
(2)$\rightarrow$(3)$\rightarrow$(4): the typer reads the rendered canvas, to decide \emph{what} happened; the localizer reads that type and the serialized trajectory records to recover
\emph{who}, \emph{where}, and \emph{when}; and the verifier recomputes every
claim from the raw trajectories as a deterministic program.}
\Description{A schematic of the TrajMind framework. On the left, a road
network and trajectory records pose the question of what is happening, and a
baseline detector emits only an anomaly score of 0.87. On the right, a frozen
vision-language model switches among LoRA adapters to act as four roles: a
screener that flags a window in one text pass, a typer that reads a rendered
canvas to name the anomaly type, a localizer that reads the serialized trajectory records
to name the participants, segment and time window, and a verifier that
recomputes the evidence and stamps the claim verified or rejected.}
\label{fig:framework}
\vskip -0.1in
\end{figure*}
In this section, we elaborate on TrajMind and figure~\ref{fig:framework} provides an overview of this framework.  Given a window of co-present
agents, TrajMind uses two inference paths.  A text-only LoRA screener monitors
every window in one pass (TrajMind$_{\mathrm{fast}}$).  For diagnosis, the slow path (TrajMind$_{\mathrm{slow}}$) applies a
canvas-based anomaly typer, a type-conditioned trajectory localizer, and an
executable verifier.  This division assigns each role the representation suited to its objective.  We first formalize the diagnosis and localization task and then describe the role-specialized adaptation, training objectives, and fast and slow inference paths.

\subsection{Problem Formulation}


Let $\mathcal{G}=(\mathcal{V},\mathcal{E})$ be a directed road graph and let
$\mathcal{W}=\{\tau_i\}_{i=1}^{N}$ denote a window of $N$ co-present agents.
Each map-matched trajectory is an ordered sequence
\[
    \tau_i=\bigl[(e_{ij},a_{ij},d_{ij})\bigr]_{j=1}^{L_i},
\]
where $e_{ij}\in\mathcal{E}$ is a traversed road segment, $a_{ij}$ is its
entry time, and $d_{ij}$ is the corresponding pass time.  We use the shared
origin $t_0=\min_{i,j}a_{ij}$ and express all predicted times relative to
$t_0$.  This makes the output invariant to absolute clock time while retaining
the ordering and duration signals 
For Porto, $e_{ij}$ and the affected region below index 100\,m grid cells,
which occupy the same segment-identifier fields in our unified data interface.

\mbox{The window label belongs to}
$\mathcal{Y}=\{\texttt{normal},\allowbreak \texttt{anomaly\_bunching},\allowbreak
\texttt{collective\_slowdown}\}$. 
For an anomalous window, the desired localized record is
\[
    z=\bigl(y,\mathcal{S},\mathcal{R},[b,e],p\bigr),
\]
where $y\in\mathcal{Y}\setminus\{\texttt{normal}\}$ is the anomaly type,
$\mathcal{S}$ is the set of participating agent identifiers,
$\mathcal{R}\subseteq\mathcal{E}$ is the affected road region, $[b,e]$ is the
event interval in seconds from $t_0$, and $p\in[0,1]$ is model confidence.  A
normal prediction contains only its type and has empty localization fields.  The
learning problem therefore couples window-level discrimination with set-valued
participant and segment recovery and temporal localization.  At inference, a
diagnosis is actionable only if these model-proposed fields can be recomputed
from $\mathcal{W}$ by the executable verifier.
Accordingly, we use \emph{alert} for the unverified fast-path output and reserve
\emph{diagnosis} for a slow-path record that passes executable verification.

\subsection{Proposed Approach}


TrajMind realizes the two operating modes in Figure~\ref{fig:framework} with a
frozen vision--language backbone and three separately trained, role-specialized
LoRA adapters.  The fast path activates LoRA-C once on a serialized trajectory
window to jointly screen and localize an event.  The slow path first activates
LoRA-A on a diagnostic canvas to determine \emph{what} happened, then switches
to LoRA-B on the serialized trajectories to recover \emph{who}, \emph{where}, and
\emph{when}. Finally, a verifier audits the proposed record.  This
decomposition lets the fast path avoid rendering, while
the slow path spends additional computation only when detailed diagnosis is needed.

\subsubsection{Shared Backbone and Role-Specialized Adaptation}

The three roles share the same frozen backbone within a model configuration,
but do not share adapter parameters.  For a selected linear projection
$W\in\mathbb{R}^{d_{\mathrm{out}}\times d_{\mathrm{in}}}$, role
$k\in\{\mathrm{A},\mathrm{B},\mathrm{C}\}$ uses the low-rank update
\citep{hu2022lora}
\[
    W_k = W + \frac{\alpha}{r}P_kQ_k,
    \qquad
    P_k\in\mathbb{R}^{d_{\mathrm{out}}\times r},\quad
    Q_k\in\mathbb{R}^{r\times d_{\mathrm{in}}}.
\]
All reported adapters for TrajMind use rank $r=16$ and scaling $\alpha=32$.  The original backbone parameters remain frozen. 
Thus, switching roles changes a small parameter
set.


Each adapter is first optimized with teacher-forced supervised fine-tuning.  If
$\mathcal{D}_k=\{(x_n^{(k)},z_n^{(k)})\}$ is the role-specific dataset and
$\phi_k=\{P_k,Q_k\}$ denotes its trainable parameters, the objective is
\[
    \mathcal{L}_{\mathrm{SFT}}^{(k)}
    =-\sum_{(x,z)\in\mathcal{D}_k}\sum_{t=1}^{|z|}
      \log p_{\theta,\phi_k}\!\left(z_t\mid x,z_{<t}\right),
\]
with frozen backbone parameters $\theta$.  The datasets and response schemas
are intentionally different: LoRA-A learns a three-way type decision from
canvases, LoRA-B learns type-conditioned localization from anomalous text windows,
and LoRA-C learns the unconditioned joint contract from both normal and anomalous
text windows.  LoRA-B is trained only on windows whose ground-truth localization is
confirmable by the verifier, preventing the localizer from being taught
claims that the deployed audit program cannot verify.  LoRA-C receives a
second, group-relative policy-optimization stage described below.

\subsubsection{Complementary\allowbreak{} Window Representations}

Raw records preserve exact identities and times, but leave population-level
structure implicit in a long sequence.  A global canvas makes
that structure visually explicit, but necessarily compresses fine-grained
identifiers and timing.  TrajMind assigns the representation according to the
role instead of forcing one encoding to serve both tasks.

The canvas encoder $\mathsf{C}(\mathcal{W})$ deterministically renders four
diagnostic panels: (A) all trajectories over the road-network layout together
with shared-segment co-occurrence, (B) mean pass time on shared segments, (C) a
three-frame temporal storyboard, and (D) entry-time strips for the busiest
shared segments.  Panels A--C expose spatial overlap, traversal delay, and
temporal evolution for type recognition, while Panel D makes headway collapse
directly visible.  The renderer contains no learned parameters and no anomaly
label. The canvas instance is shown in case study (section~\ref{sec:case-study-compact}).

The text encoder $\mathsf{T}(\mathcal{W})$ retains the map-matched records in two
linked views.  A shared-segment table lists co-present counts, relative entry
times, and pass times, followed by each agent's ordered
\texttt{segment:entry/pass} trajectory.  LoRA-B and LoRA-C consequently emit original
agent and segment identifiers rather than canvas handles, and their
\texttt{start\_s} and \texttt{end\_s} fields use the same $t_0$ as the serialized
input.  This common contract allows the verifier to map every generated field
back to the raw window.


\subsubsection{TrajMind$_{\mathrm{slow}}$: Type, Localize, and Verify}

The slow path factorizes diagnosis because the information useful for choosing
an anomaly type is not identical to that needed for exact localization. Given a
window, LoRA-A performs canvas-based typing,
\[
    \hat y=\arg\max_{y\in\mathcal{Y}}
    p_{\theta,\phi_{\mathrm{A}}}\!\left(y\mid
    \mathsf{C}(\mathcal{W})\right),
\]
and emits only \texttt{\{"type": $\hat y$\}}.  Training includes normal windows
and both collective anomaly types, so this adapter owns the anomaly gate rather
than receiving an oracle-positive window.  If $\hat y=\texttt{normal}$, the
slow path terminates and produces no event hypothesis.

For a non-normal decision, LoRA-B receives the predicted type as an explicit
condition and reads the text representation:
\[
    \hat g=(\hat{\mathcal{S}},\hat{\mathcal{R}},
    [\hat b,\hat e],\hat p)
    \sim p_{\theta,\phi_{\mathrm{B}}}\!\left(
    g\mid\mathsf{T}(\mathcal{W}),\hat y\right).
\]
During SFT, the condition is the labeled anomaly type; at deployment it is
LoRA-A's output.  The response schema fixes the task boundary: LoRA-B predicts
participants, segments, relative start and end times, and confidence, but is
not allowed to revise the type.  

\paragraph{Executable verification.}
Following work that grounds model reasoning through external actions and tool
feedback~\citep{yao2023react,gou2024critic}, the verifier treats
$(\hat y,\hat g)$ as a hypothesis, never as evidence.  It
first rejects invalid identifiers and checks that the named agents are
co-present and traverse the named region.  For a slowdown claim, each named
agent's largest standardized delay on the region is computed from historical
segment statistics,
\[
    z_i=\max_{e\in\hat{\mathcal{R}}\cap\tau_i}
    \frac{d_{i,e}-\mu_e}{\sigma_e},
\]
and the verifier measures the fraction of participants above the calibrated
slowdown threshold.  For a bunching claim, it sorts the participants' entry
times on the first named segment and computes
\[
    h=\frac{a_{(m)}-a_{(1)}}{m-1},
\]
where $m$ is the number of named agents observed on that segment; a small $h$
supports collapsed headway.  These type-specific checks are accompanied by
passed-through and regional-density checks, yielding for every tool $j$ a
support--contradiction--uncertainty tuple $(s_j,c_j,u_j)$ and its measured
evidence.

Let $w_j^{(\hat y)}$ denote the type-specific tool weight and
$q_j=\max(s_j,c_j)$ its informativeness.  Evidence is aggregated as
\[
    s=\frac{\sum_j w_j^{(\hat y)}q_js_j}
            {\sum_j w_j^{(\hat y)}q_j},
    \qquad
    c=\max_j c_j,
    \qquad
    u=\max\{0,1-\max(s,c)\}.
\]
Slowdown assigns weights $0.5/0.2/0.3$ to the slowdown,
passed-through, and density checks; bunching uses $0.6/0.2/0.2$ for the
corresponding three checks.  A single strong contradiction can therefore veto
several weakly supportive measurements.  A hypothesis is released only when
$s\geq0.70$ and $c<0.65$; $c\geq0.65$ rejects it, and an intermediate case is
withheld.  For the retained record, model confidence is
recalibrated by
\[
    p_{\mathrm{final}}=\sigma\!\left(
    \operatorname{logit}(\hat p)+s-1.5c-0.5u\right).
\]
The retained diagnosis contains the raw measurements of each executed check in
addition to $(s,c,u)$, so the slow-path output is supported by recomputed
evidence.

\subsubsection{TrajMind$_{\mathrm{fast}}$: Single-Pass Monitoring}

The slow factorization improves auditability but incurs rendering, two model
passes, and symbolic verification.  LoRA-C removes these dependencies from the
latency-critical path.  Only based on text representation, it decodes
\[
    \hat z_{\mathrm{fast}}
    \sim p_{\theta,\phi_{\mathrm{C}}}\!\left(z\mid
    \mathsf{T}(\mathcal{W})\right).
\]
A normal window returns only \texttt{\{"type":"normal"\}}, while an anomalous
window returns type, participants, segments, relative interval, and confidence
in one JSON object.  The real-time interface foregrounds the anomaly flag and
participant list shown in Figure~\ref{fig:framework}, while retaining the other
structured fields for evaluation and downstream triage.  

Pure SFT teaches the output grammar and the joint mapping, but it does not
directly optimize the asymmetric costs of hallucinating events, missing true
events, and partially localizing a correct decision.  We therefore continue
LoRA-C with Group Relative Policy Optimization (GRPO)
\citep{shao2024deepseekmath}. Recent work improves the efficiency or
informativeness of agentic post-training through online-verified prompt
selection, auxiliary world-model supervision from policy rollouts, and
cooperative parameter-subspace evolution
\citep{wu2026movingedge,lu2026policyworld,wang2026copes}. These directions are
complementary to our structured reward: TrajMind keeps the training procedure
simple and optimizes the application-specific detection and localization
contract. Let $F_{\mathcal{S}}$ and $F_{\mathcal{R}}$ be
set F1 for participants and segments, respectively, let $I_t$ be temporal IoU,
and define $L=F_{\mathcal{S}}+F_{\mathcal{R}}+0.5I_t$.  The scalar reward is designed as
\[
R=\begin{cases}
-1, & \text{invalid output},\\
0.3+1, & y=\hat y=\texttt{normal},\\
0.3+0.5s-0.8c,
    & y=\texttt{normal},\ \hat y\neq\texttt{normal},\\
0.3, & y\neq\texttt{normal},\ \hat y=\texttt{normal},\\
0.3+1+L, & y=\hat y\neq\texttt{normal},\\
0.3+0.2+0.25L,
    & y,\hat y\neq\texttt{normal},\ y\neq\hat y.
\end{cases}
\]
Here the $0.3$ term rewards a valid schema.  On true anomalies, localization
credit is largest for a correct type but remains weakly informative after a
wrong non-normal type, while abstaining receives no detection or localization credit.
On normal windows, a hallucinated anomaly is additionally scored by verifier
support $s$ and contradiction $c$ for the named hypothesis.  The verifier term
is deliberately not used on true-anomaly rewards, so LoRA-C does not learn to
imitate the verifier's finite-coverage thresholds.

For each prompt $i$, GRPO samples a group of completions $j$ and normalizes their
rewards into
\[
    \widehat A_{ij}=\frac{R_{ij}-\operatorname{mean}_j R_{ij}}
    {\operatorname{std}_j R_{ij}+\epsilon}.
\]
The policy update uses these within-prompt relative advantages with eight
completions per group, temperature $1.15$, and no KL penalty ($\beta=0$).  This
stage updates only $\phi_{\mathrm{C}}$, the parameter of adapter C.  Consequently, the deployed
system is a switchable family of compact role parameters: LoRA-C serves the
always-on fast monitor, whereas an escalated window swaps in LoRA-A and then
LoRA-B before deterministic verification.

\subsubsection{Inference Procedure}

Algorithm~\ref{alg:trajmind-procedures} summarizes the two inference modes.  The
trained adapters and frozen backbone remain fixed throughout inference.  We
write $(v,\Xi)=\mathsf{Verifier}(\mathcal{W},\eta)$ for the verifier verdict and
its supporting evidence for hypothesis $\eta$.  The fast path returns after one
decode; the slow path releases a diagnosis only after localization and
verification.

\begin{algorithm}[H]
\caption{TrajMind's dual-path inference pipeline.}
\label{alg:trajmind-procedures}
\footnotesize
\begin{algorithmic}[1]
\State \textbf{Input:} A trajectory window $\mathcal{W}$ and mode $\rho\in\{\mathrm{fast},\mathrm{slow}\}$
\State \textbf{Output:} A fast-path prediction or a slow-path diagnosis verdict
\If{$\rho=\mathrm{fast}$}
    \State $\hat z_{\mathrm{fast}}\gets\Call{Screener}{\mathsf{T}(\mathcal{W})}$
    \State \Return $\hat z_{\mathrm{fast}}$ \Comment{normal verdict or structured alert}
\EndIf
\State $\hat y\gets\Call{Typer}{\mathsf{C}(\mathcal{W})}$
\If{$\hat y=\texttt{normal}$}
    \State \Return no diagnosis
\EndIf
\State $\hat g\gets\Call{Localizer}{\mathsf{T}(\mathcal{W}),\hat y}$
\State $(v,\Xi)\gets\Call{Verifier}{\mathcal{W},(\hat y,\hat g)}$
\If{$v=\textsc{verified}$}
    \State \Return diagnosis $(\hat y,\hat g,\Xi)$
\EndIf
\State \Return $v$ \Comment{rejected or withheld}
\end{algorithmic}
\end{algorithm}

\section{Experiments}

We aim to answer the following research questions:

\begingroup\small\emergencystretch=1em
\noindent\textbf{RQ1:} How good TrajMind is at collective anomaly
detection and typing?\par
\noindent\textbf{RQ2:} How accurately does TrajMind localize the
participants, road segments, and time of a collective anomaly?\par
\noindent\textbf{RQ3:} Is TrajMind robust to unseen cities and
unseen anomaly strengths?\par
\noindent\textbf{RQ4:} How efficient is TrajMind$_{\mathrm{fast}}$,
and what does it trade away?\par
\noindent\textbf{RQ5:} How do role specialization and RL refinement affect
TrajMind?\par
\noindent\textbf{RQ6:} Is the zero-shot ceiling of frozen VLMs a perception
limit or a calibration limit?\par
\endgroup

\subsection{Experimental Settings}
\label{sec:experimental-settings}

In this section, we briefly introduce the datasets, tasks, evaluation
metrics, and baselines.  

\textbf{Datasets.} We use Chengdu and Xi'an from DiDi GAIA~\cite{didi2017gaia}
and Porto from the ECML/PKDD-2015 challenge~\cite{moreiramatias2015porto}.
Chengdu and Xi'an retain map-matched road segments as shown in Table~\ref{tab:dataset-summary}.


\newsavebox{\datasettablebox}

\begin{table}[h]
\centering
\vskip -0.1in
\caption{Dataset statistics.}
\label{tab:dataset-summary}
\small
\setlength{\tabcolsep}{7pt}
\renewcommand{\arraystretch}{1.00}
\sbox{\datasettablebox}{%
\begin{tabular}{lcc}
\toprule
Dataset & \shortstack{Trajectories\\train/eval}
  & \shortstack{Group windows\\train/eval} \\
\midrule
Chengdu & 70k/14k & 838/189 \\
Xi'an & 70k/14k & 192/190 \\
Porto & 70k/14k & 1,451/351 \\
\bottomrule
\end{tabular}
}
\usebox{\datasettablebox}

\begin{minipage}{\wd\datasettablebox}
\footnotesize
\raggedright
Windows are retained 900-s buckets with at least 3 agents.
\end{minipage}
\vskip -0.1in
\end{table}

Because no known public dataset provides real-world collective traffic anomaly
labels, we synthetically inject anomalies into real trajectories, and
the injection record supplies exact ground truth for the type, participants,
segment, and interval: a \emph{collective slowdown} inflates each
participant's pass time on a shared segment by a factor drawn from
$\mathcal{U}[1.6, 2.6]$, and \emph{bunching} compresses their arrivals there
into $\mathcal{U}[10, 40]$\,s. Each test window is assigned a label
$1{:}1{:}1$ over \{normal, slowdown, bunching\} and injected accordingly; for
the ${\sim}5\%$ of windows where injection is infeasible, the window reverts
to normal.  Learned TrajMind components are optimized
only on Chengdu. Xi'an and Porto therefore evaluate
zero-shot transfer.  Two held-out protocols with stronger (B) and weaker (C)
amplitudes and a timing-preserving structure-only control probe robustness.

\textbf{Tasks and Evaluation Metrics.} We evaluate four tasks.  For
\emph{detection and typing}, every window receives one label from \{normal,
slowdown, bunching\}, and we report balanced accuracy (chance $.333$)
together with a binary view that merges the two anomaly classes (chance
$.500$).  For \emph{localization}, {a separate pass injects an anomaly
into every test window (slowdown:bunching $1{:}1$), and the windows where
injection succeeds form the evaluation set ($N = 179/189/347$ for
Chengdu/Xi'an/Porto) with no model- or verifier-based filtering}.
We report participant F1, segment F1, temporal IoU (tIoU), and response coverage for localization. The two F1 metrics score the participant and segment identifiers emitted by the model.
For Porto, segment identifiers denote the 100\,m grid cells described above.
Abstentions score zero on both F1 metrics but are excluded from tIoU, so tIoU
must be read jointly with coverage. For \emph{robustness}, the same metrics are
reported under city shift and protocols B/C, plus the false-positive rate on
the structure-only control. For \emph{efficiency}, we measure wall-clock
latency per window on one NVIDIA H20 at batch size one.  

\textbf{Baselines.} We compare against five kinds of methods: (i) a
training-free \emph{statistical temporal rule} over held-out per-segment
pass-time statistics; (ii) the symbolic localizer \emph{Derive} and the
hybrid \emph{Cascade}, which falls back to a VLM only where Derive abstains;
(iii) \emph{DSAB}~\cite{hu2023dsab}, the published learned detector whose
task unit matches ours, adapted to our data and reported in the binary panel
only since it emits no type; (iv) \emph{Traj-MLLM}~\cite{liu2025trajmllm},
the closest trajectory-centric VLM, adapted from per-trajectory
classification to group windows; and (v) \emph{zero-shot prompting} of frozen
VLMs on both the text serialization and the canvas rendering for four
open-weight backbones (Qwen3-VL-2B/8B~\cite{bai2025qwen3vl} and
Qwen2.5-VL-3B/7B~\cite{bai2025qwen25vl}), and on the canvas alone for three
API models (Qwen3-VL-Plus, Qwen3-VL-235B-A22B, and
GLM-5V-Turbo\footnote{No technical report
for \texttt{glm-5v-turbo}, so we cite the most recent report of GLM~\cite{glmv2025glm41v}.}).  Both TrajMind paths are LoRA
adapters~\cite{hu2022lora} over the same frozen backbones, and
TrajMind$_{\mathrm{fast}}$ is additionally refined from its supervised
checkpoint with GRPO~\cite{shao2024deepseekmath}.
TrajMind$_{\mathrm{slow}}$'s typing adapter is reported at all four
open-weight scales; all other learned arms use Qwen2.5-VL-3B-Instruct.

%
%

\begin{table}[tb]
\centering
\caption{Detection and anomaly-typing performance. \textbf{Bold} marks the best reported
performance (balanced accuracy) and \underline{underlined} values mark the strongest
baseline.}
\label{tab:t6-comprehensive}
\small
\setlength{\tabcolsep}{3.5pt}
\renewcommand{\arraystretch}{0.98}
\begin{tabular}{>{\raggedright\arraybackslash}p{0.48\columnwidth}ccc}
\toprule
Method and setting & Chengdu & Xi'an & Porto \\
\midrule
\rowcolor{bgTableSection}
\multicolumn{4}{l}{\textit{Three-way anomaly typing}} \\
\quad Statistical temporal rule & .573 & .535 & .533 \\
\quad Traj-MLLM$^{b}$
  & .523 & .415 & .423 \\
\specialrule{0.35pt}{0.35pt}{0.35pt}
\multicolumn{4}{l}{\quad Text-only zero-shot} \\
\hspace{2em}Qwen3-VL-2B-Instruct & .333 & .333 & .333 \\
\hspace{2em}Qwen2.5-VL-3B-Instruct & .330 & .335 & .333 \\
\hspace{2em}Qwen2.5-VL-7B-Instruct & .356 & .369 & .355 \\
\hspace{2em}Qwen3-VL-8B-Instruct & .339 & .351 & .304 \\
\specialrule{0.35pt}{0.35pt}{0.35pt}
\multicolumn{4}{l}{\quad Canvas zero-shot: open-source} \\
\hspace{2em}Qwen3-VL-2B-Instruct & .333 & .333 & .333 \\
\hspace{2em}Qwen2.5-VL-3B-Instruct & .333 & .333 & .333 \\
\hspace{2em}Qwen2.5-VL-7B-Instruct & .522 & .681 & \underline{.774} \\
\hspace{2em}Qwen3-VL-8B-Instruct & \underline{.670} & \underline{.691} & .669 \\
\specialrule{0.35pt}{0.35pt}{0.35pt}
\multicolumn{4}{l}{\quad Canvas zero-shot: API} \\
\hspace{2em}Qwen3-VL-Plus & .540 & .420 & .430 \\
\hspace{2em}GLM-5V-Turbo & .574 & .521 & .540 \\
\hspace{2em}Qwen3-VL-235B-A22B & .667 & .670 & .667 \\
\specialrule{0.35pt}{0.35pt}{0.35pt}
\rowcolor{bgTrajMind}
\quad TrajMind$_{\mathrm{fast}}$ (3B)
  & .637 & .626 & .614 \\
\rowcolor{bgTrajMind}
\multicolumn{4}{l}{\quad TrajMind$_{\mathrm{slow}}$} \\
\rowcolor{bgTrajMind}
\hspace{2em}Qwen3-VL-2B-Instruct
  & .952 & .942 & .914 \\
\rowcolor{bgTrajMind}
\hspace{2em}Qwen2.5-VL-3B-Instruct
  & .960 & .954 & .927 \\
\rowcolor{bgTrajMind}
\hspace{2em}Qwen2.5-VL-7B-Instruct
  & .972 & .966 & .945 \\
\rowcolor{bgTrajMind}
\hspace{2em}Qwen3-VL-8B-Instruct
  & \textbf{.983} & \textbf{.980} & \textbf{.959} \\
\midrule
\rowcolor{bgTableSection}
\multicolumn{4}{l}{\textit{Binary anomaly detection}$^{c}$} \\
\quad DSAB$^{a}$ & \underline{.680} & \underline{.725} & \underline{.552} \\
\specialrule{0.35pt}{0.35pt}{0.35pt}
\rowcolor{bgTrajMind}
\quad TrajMind$_{\mathrm{fast}}$ (3B)
  & .960 & .949 & .935 \\
\rowcolor{bgTrajMind}
\multicolumn{4}{l}{\quad TrajMind$_{\mathrm{slow}}$} \\
\rowcolor{bgTrajMind}
\hspace{2em}Qwen3-VL-2B-Instruct
  & .968 & .969 & .955 \\
\rowcolor{bgTrajMind}
\hspace{2em}Qwen2.5-VL-3B-Instruct
  & .974 & .971 & .963 \\
\rowcolor{bgTrajMind}
\hspace{2em}Qwen2.5-VL-7B-Instruct
  & .984 & .981 & .975 \\
\rowcolor{bgTrajMind}
\hspace{2em}Qwen3-VL-8B-Instruct
  & \textbf{.987} & \textbf{.989} & \textbf{.977} \\
\bottomrule
\end{tabular}

\vspace{2pt}
\begin{minipage}{\columnwidth}
\footnotesize
\(^{a}\)DSAB is a binary-only graph baseline adapted from individual anomaly detection, and it provides no anomaly type.
\(^{b}\)This is also adapted from individual anomaly detection and Qwen3-VL-Plus is the backbone.
\(^{c}\)The three-way label space is \{normal, collective slowdown, bunching\}
(chance \(.333\)); the binary label space merges the two anomaly types (chance
\(.500\)). TrajMind$_{\mathrm{slow}}$ uses a canvas typing adapter, whereas
TrajMind$_{\mathrm{fast}}$ reports the single-pass text adapter.
\end{minipage}
\vskip -0.15in
\end{table}

\begingroup
\def\LocalizationTableOnly{}
\ifdefined\LocalizationProseOnly
\else
\begin{table*}[t]
\centering
\caption{Unified localization comparison: participant F1 (subject F1),
temporal IoU (tIoU), model-emitted segment F1, and response coverage across
Chengdu (CD), Xi'an (XA), and Porto (PT).  Coverage is the fraction of windows for which a method returns an
answer. \textbf{Bold} marks the best,
\underline{underlining} marks the second-best, and
* marks the strongest non-TrajMind baseline.} 
\label{tab:t7-localization}
\small
\setlength{\tabcolsep}{3pt}
\renewcommand{\arraystretch}{0.90}
\begin{tabularx}{\textwidth}{>{\raggedright\arraybackslash}p{0.225\textwidth}*{12}{Y}}
\toprule
& \multicolumn{3}{c}{Subject F1} & \multicolumn{3}{c}{tIoU}
& \multicolumn{3}{c}{Segment F1} & \multicolumn{3}{c}{Response coverage} \\
\cmidrule(lr){2-4}\cmidrule(lr){5-7}\cmidrule(lr){8-10}\cmidrule(lr){11-13}
Method & CD & XA & PT & CD & XA & PT
       & CD & XA & PT & CD & XA & PT \\
\midrule
\quad Derive (symbolic heuristic)
  & .640 & .526 & .510 & n/a & n/a & n/a
  & .337 & .346 & .500 & .727 & .648 & .536 \\
\quad Cascade (Derive $\rightarrow$ VLM)
  & .761\textsuperscript{*} & .794\textsuperscript{*} & \underline{.803}\textsuperscript{*} & .019 & .019 & .024
  & .367\textsuperscript{*} & .498\textsuperscript{*} & \underline{.746}\textsuperscript{*}
  & \underline{.991} & .988 & .979 \\
\quad Traj-MLLM
  & .396 & .395 & .471 & .220\textsuperscript{*} & .234\textsuperscript{*} & .240\textsuperscript{*}
  & .000 & .000 & .000 & .967 & {.992} & .987 \\
\specialrule{0.35pt}{0.35pt}{0.35pt}
\multicolumn{13}{l}{\quad Canvas zero-shot: open-source} \\
\hspace{2em}Qwen3-VL-2B-Instruct
  & .108 & .158 & .156 & .009 & .013 & .012
  & .009 & .058 & .092 & .886 & .719 & .795 \\
\hspace{2em}Qwen2.5-VL-3B-Instruct
  & .113 & .173 & .171 & .031 & .028 & .031
  & .002 & .057 & .122 & .955 & .954 & .958 \\
\hspace{2em}Qwen2.5-VL-7B-Instruct
  & .209 & .314 & .301 & .037 & .038 & .046
  & .047 & .196 & .221 & .940 & .910 & .886 \\
\hspace{2em}Qwen3-VL-8B-Instruct
  & .102 & .115 & .132 & .026 & .039 & .040
  & .006 & .042 & .088 & .560 & .457 & .502 \\
\specialrule{0.35pt}{0.35pt}{0.35pt}
\multicolumn{13}{l}{\quad Canvas zero-shot: API} \\
\hspace{2em}Qwen3-VL-Plus
  & .345 & .435 & .448 & .026 & .029 & .021
  & .003 & .002 & .005
  & \textbf{1.000}\textsuperscript{*} & \textbf{1.000}\textsuperscript{*} & \textbf{1.000}\textsuperscript{*} \\
\hspace{2em}GLM-5V-Turbo
  & .368 & .498 & .498 & .035 & .042 & .040
  & .010 & .024 & .035 & .979 & .959 & .970 \\
\hspace{2em}Qwen3-VL-235B-A22B
  & .132 & .220 & .199 & .029 & .032 & .039
  & .011 & .022 & .020
  & \textbf{1.000}\textsuperscript{*} & \underline{.998} & \underline{.999} \\
\specialrule{0.35pt}{0.35pt}{0.35pt}
\rowcolor{bgTrajMind}
\quad TrajMind$_{\mathrm{fast}}$ (3B)
  & \underline{.950} & \underline{.972} & .722
  & \textbf{.791} & \underline{.787} & \textbf{.717}
  & \textbf{.609} & \underline{.555} & .597 & .951 & .977 & .730 \\
\rowcolor{bgTrajMind}
\quad TrajMind$_{\mathrm{slow}}$ (3B)
  & \textbf{.993} & \textbf{.995} & \textbf{.941}
  & \underline{.753} & \textbf{.796} & {\underline{.379}}
  & \underline{.502} & \textbf{.636} & \textbf{.930}
  & \textbf{1.000} & \textbf{1.000} & \textbf{1.000} \\
\bottomrule
\end{tabularx}

\vspace{2pt}
\begin{minipage}{\textwidth}
\scriptsize
\setlength{\leftmargini}{1.5em}
\begin{itemize}
  \setlength{\itemsep}{0pt}
  \setlength{\parsep}{0pt}
  \setlength{\parskip}{0pt}
  \setlength{\topsep}{1pt}
  \item \textbf{Metrics.} Higher is better; tIoU is averaged over answered cases and should be read with response coverage; Derive emits no interval, so its tIoU is undefined and marked ``n/a''.
  \item \textbf{Traj-MLLM.} It uses Qwen3-VL-Plus with group adaptation through
  ten window renderings.
\end{itemize}
\end{minipage}
\end{table*}
\fi

\ifdefined\LocalizationTableOnly
\else
\subsection{Fine-Grained Localization (RQ2, Table~\ref{tab:t7-localization})}
\label{sec:t7-localization}

To answer \textbf{RQ2}, we compare TrajMind with symbolic and hybrid localizers,
Traj-MLLM, and frozen canvas-prompted VLMs.  \textbf{Table~\ref{tab:t7-localization}}
yields three observations.

\textbf{First, the type-conditioned slow specialist provides the most reliable
participant localization.}  TrajMind$_{\mathrm{slow}}$ reaches subject F1 scores
of $.993/.995/.941$ on Chengdu, Xi'an, and Porto, outperforming the strongest
non-TrajMind baseline by $.232/.201/.138$, respectively.  It also achieves full
coverage.  The best symbolic hybrid covers $.991/.988/.979$, whereas the fast
path covers $.951/.977/.730$.  
Under this localization protocol, the slow path receives the ground-
truth anomaly type, while the fast path types and localizes in one pass; the
comparison therefore isolates the benefit of type-conditioned, lossless-text
localization rather than claiming identical inference contracts.

\textbf{Second, the two specialists are complementary across localization
dimensions.}  The fast path obtains the highest tIoU on Chengdu and Porto
($.791$ and $.717$) and is only $.009$ behind the slow path on Xi'an.  It also
leads segment F1 on Chengdu ($.609$).  In contrast, the slow path is strongest
for segment localization after transfer, reaching $.636$ on Xi'an and $.930$ on
Porto, gains of $.081$ and $.333$ over the fast path. {Its lower Porto tIoU ($.379$) shows that accurate participants and segments do not automatically
fix temporal boundaries.}  Conversely, the fast path's $.717$ Porto tIoU is
averaged only over the $.730$ of windows it answers.  Reading tIoU jointly with
coverage therefore reveals a precision--coverage trade-off, rather than
uniform temporal superiority by either path.

\textbf{Third, fine-grained localization requires task specialization, not merely
larger models or more frequent answers.} The frozen API VLMs respond on at
least $.959$ of windows, yet their segment F1 remains between $.002$ and $.035$;
similarly, Traj-MLLM covers $.967$--$.992$ but obtains zero segment F1 in all
three cities.  Scaling is also non-monotonic: within Qwen2.5-VL, moving from
3B to 7B improves subject F1 in every city, whereas the Qwen3-VL 8B model is
worse than its 2B counterpart and abstains on roughly half the windows.  Thus,
response rate and parameter count do not substitute for a localization
contract aligned with agent identities and trajectory structure.  The large
gap between the generic baselines and both TrajMind specialists attributes the
gain primarily to role-specific adaptation.
\fi

\endgroup

\subsection{Anomaly Detection/Typing (RQ1, Table~\ref{tab:t6-comprehensive})}
\label{sec:t6-comprehensive}
\enlargethispage{1pt}

To answer \textbf{RQ1}, we compare TrajMind with statistical, graph-based, and
trajectory-MLLM baselines, as well as text-only and canvas-based zero-shot
VLMs.  Results in \textbf{table~\ref{tab:t6-comprehensive}} yields three observations.

\textbf{First, role-specialized adaptation is consistently more effective than
the competing detection and typing paradigms.}  TrajMind$_{\mathrm{slow}}$
achieves the best three-way balanced accuracy in every city. {With the 8B
backbone, it reaches \(.983\), \(.980\), and \(.959\) on Chengdu, Xi'an, and
Porto, exceeding the strongest non-TrajMind result in each city by
\(.313\), \(.289\), and \(.185\), respectively.}  The advantage does not rely
on a large backbone: even the 2B specialist obtains
\(.952/.942/.914\), remaining above all baselines.  The same ordering holds
for binary detection, where the 8B slow specialist improves over DSAB by
\(.307/.264/.425\).  These consistent gains indicate that adapting the model
to the group-level traffic decision is more consequential than relying on a
generic anomaly score or an individually oriented trajectory MLLM.

\textbf{Second, the input representation determines whether zero-shot scaling
is useful.}  Text-only prompting remains close to the \(.333\) chance level
for all four local backbones and all three cities.  Rendering the same task as
a diagnostic canvas produces a clear scale-dependent improvement: the 7B and
8B models reach as high as \(.774\), whereas the 2B and 3B models still
collapse to chance.  Nevertheless, scale alone does not close the gap.  The
235B API model records only \(.667/.670/.667\), comparable to the local 8B
model and substantially below every adapted slow specialist.  Thus, the
canvas exposes the collective spatiotemporal pattern to sufficiently capable
VLMs, while traffic-specific adaptation is still required to place a reliable
decision boundary.

\textbf{Third, detection and fine-grained typing are distinct capabilities.}
TrajMind$_{\mathrm{fast}}$ attains \(.960/.949/.935\) on binary detection,
outperforming DSAB by \(.280/.224/.383\), but its corresponding three-way
scores are \(.637/.626/.614\).  In contrast,
TrajMind$_{\mathrm{slow}}$ remains near ceiling in both panels.  Together,
these results expose a clear division of labor.  The fast
specialist therefore preserves strong anomaly screening while giving up partial semantic resolution needed to separate slowdown from bunching. However, the
canvas specialist supplies the resolution.  This result supports treating
the two paths as complementary operating specialists.


\begingroup
\def\LocalizationProseOnly{}

\endgroup

%
%
\subsection{Cross-City Localization (\textbf{RQ3}, Table~\ref{tab:cross-city-localization}) }
\label{sec:cross-city-localization}

\begin{table}[tb]
\centering
\caption{Cross-city localization performance of TrajMind, trained on Chengdu and tested on Xi'an and Porto.  \textbf{Bold} marks the best performance.}
\label{tab:cross-city-localization}
\small
\setlength{\tabcolsep}{6pt}
\renewcommand{\arraystretch}{0.96}
\begin{tabular}{lcccc}
\toprule
& \multicolumn{2}{c}{Xi'an} & \multicolumn{2}{c}{Porto} \\
\cmidrule(lr){2-3}\cmidrule(lr){4-5}
Metric & Fast & Slow & Fast & Slow \\
\midrule
Subject F1   & .972 & \textbf{.995} & .722 & \textbf{.941} \\
tIoU         & .787 & \textbf{.796} & \textbf{.717} & {.379} \\
Segment F1   & .555 & \textbf{.636} & .597 & \textbf{.930} \\
Coverage     & .977 & \textbf{1.000} & .730 & \textbf{1.000} \\
\bottomrule
\end{tabular}
\end{table}

To answer \textbf{RQ3}, we apply the two Chengdu-trained localization specialists to
Xi'an and Porto without target-city adaptation.  \textbf{Table~\ref{tab:cross-city-localization}}
yields three observations.

\textbf{First, the slow path preserves reliable participant localization across
both unseen cities.}  It reaches subject F1 scores of $.995$ on Xi'an and
$.941$ on Porto, improving over the fast path by $.023$ and $.219$,
respectively.  The paths are close on Xi'an, whereas the wider Porto gap shows
that the type-conditioned, lossless-text localizer is more robust under this
target-city shift.  The slow path is evaluated with the ground-truth anomaly
type and a forced response, so its $1.000$ coverage reflects its diagnostic
contract rather than an isolated localization gain.  Nevertheless, the high
subject F1 shows that its returned participants remain
accurate after transfer.

\textbf{Second, spatial localization transfers even when the target spatial
representation changes.}  The slow path improves segment F1 over the fast
path by $.081$ on Xi'an and $.333$ on Porto, reaching $.930$ on Porto despite
that corpus using 100\,m grid cells in place of the road-segment identifiers
available in Chengdu and Xi'an.  The Porto score therefore evaluates whether
the transferred model directly emits the correct grid-cell identifiers, rather
than inferring location from participant predictions after generation.

\textbf{Third, temporal localization remains city dependent and exposes a
coverage--precision trade-off.}  On Xi'an, the two paths obtain nearly
identical tIoU ($.787$ versus $.796$).  On Porto, the fast path has the higher
tIoU ($.717$ versus $.379$), but its score is computed only on the $73.0\%$ of
windows for which it returns an answer; the slow path returns an interval for
all windows.  The Porto tIoU gap therefore does not establish end-to-end
temporal superiority for the fast path.  Rather, it shows that complete
cross-city localization does not by itself guarantee accurate temporal
boundaries.

\newif\ifFastAblationFigureOnly
\FastAblationFigureOnlytrue
\ifFastAblationFigureOnly


\definecolor{fastSFT}{HTML}{7699C7}
\definecolor{fastGRPO}{HTML}{C85C5C}
\pgfplotsset{
  fast ablation plot/.style={
    ybar=1.2pt,
    bar width=10pt,
    width=\linewidth,
    height=4.25cm,
    ymin=0,
    ymax=1.10,
    symbolic x coords={Chengdu,Xi'an,Porto},
    xtick=data,
    enlarge x limits=0.20,
    ytick={0,.2,.4,.6,.8,1.0},
    yticklabel style={font=\footnotesize},
    xticklabel style={font=\footnotesize},
    tick align=outside,
    tick style={black!60},
    axis line style={black!70},
    axis background/.style={fill=white},
    grid=major,
    grid style={draw=black!10},
    title style={font=\small\bfseries},
    ylabel style={font=\footnotesize},
    scaled y ticks=false,
    nodes near coords={\pgfmathprintnumber[fixed,precision=2,fixed zerofill]{\pgfplotspointmeta}},
    every node near coord/.append style={font=\tiny,anchor=south},
    clip=false,
  },
  fast sft bars/.style={
    fill=fastSFT!78,
    draw=fastSFT!45!black,
  },
  fast grpo bars/.style={fill=fastGRPO!30,draw=fastGRPO!50!black,
    pattern=north east lines,pattern color=fastGRPO!70!black},
  fast shortened y break/.style={
    axis y discontinuity=none,
    after end axis/.append code={
      \draw[white,line width=1.1pt]
        ([yshift=1.8pt]rel axis cs:0,0) -- ([yshift=6.2pt]rel axis cs:0,0);
      \draw[black!70,line width=.4pt,line cap=round]
        ([xshift=-1.6pt,yshift=2.2pt]rel axis cs:0,0)
        -- ([xshift=1.6pt,yshift=3.4pt]rel axis cs:0,0);
      \draw[black!70,line width=.4pt,line cap=round]
        ([xshift=-1.6pt,yshift=4.4pt]rel axis cs:0,0)
        -- ([xshift=1.6pt,yshift=5.6pt]rel axis cs:0,0);
    },
  },
  fast ablation compact/.style={
    fast ablation plot,
    ybar=2.4pt,
    bar width=8pt,
    width=\linewidth,
    height=2.75cm,
    ymax=1.18,
    enlarge x limits=0.20,
    ytick={0,.5,1.0},
    yticklabel style={font=\fontsize{6.8}{7.1}\selectfont,xshift=-2pt},
    xticklabel style={font=\fontsize{6.8}{7.1}\selectfont},
    major tick length=2pt,
    xtick align=outside,
    ytick align=inside,
    axis x line*=bottom,
    axis y line*=left,
    axis line style={black!70,-{Stealth[length=3pt,width=2.2pt]}},
    title style={font=\footnotesize\bfseries},
    every node near coord/.append style={
      font=\fontsize{6.3}{6.6}\selectfont,
      rotate=0,
      anchor=south,
      inner sep=.5pt,
    },
  },
}

\newcommand{\fastablationlegend}{%
  \tikz[baseline=-0.5ex]{\draw[fill=fastSFT!78,draw=fastSFT!45!black]
    (0,0) rectangle (0.20,0.12);}~SFT (supervised)\qquad
  \tikz[baseline=-0.5ex]{\draw[fill=fastGRPO!30,draw=fastGRPO!50!black,
    pattern=north east lines,pattern color=fastGRPO!70!black]
    (0,0) rectangle (0.20,0.12);}~GRPO (RL-refined)%
}

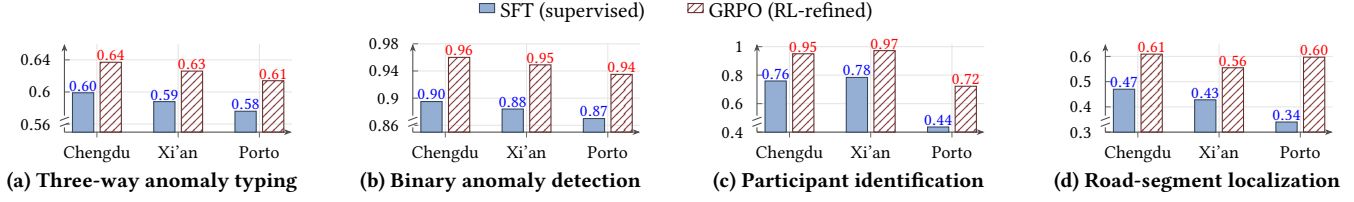
\begin{figure*}[t]
\centering
{\small\fastablationlegend}\par\vspace{2pt}
\makebox[\textwidth][c]{%
\begin{minipage}[t]{0.258\textwidth}
\centering
\begin{tikzpicture}
\begin{axis}[
  fast ablation compact,
  ymin=.55,
  ymax=.66,
  ytick={.56,.60,.64},
  fast shortened y break,
]
\addplot+[fast sft bars] coordinates
  {(Chengdu,.599) (Xi'an,.588) (Porto,.576)};
\addplot+[fast grpo bars] coordinates
  {(Chengdu,.637) (Xi'an,.626) (Porto,.614)};
\end{axis}
\end{tikzpicture}
\par\vspace{-5pt}{\small\bfseries (a) Three-way anomaly typing\strut}
\end{minipage}%
\hspace{0.001\textwidth}%
\begin{minipage}[t]{0.258\textwidth}
\centering
\begin{tikzpicture}
\begin{axis}[
  fast ablation compact,
  ymin=.85,
  ymax=.98,
  ytick={.86,.90,.94,.98},
  fast shortened y break,
]
\addplot+[fast sft bars] coordinates
  {(Chengdu,.895) (Xi'an,.884) (Porto,.870)};
\addplot+[fast grpo bars] coordinates
  {(Chengdu,.960) (Xi'an,.949) (Porto,.935)};
\end{axis}
\end{tikzpicture}
\par\vspace{-5pt}{\small\bfseries (b) Binary anomaly detection\strut}
\end{minipage}%
\hspace{0.001\textwidth}%
\begin{minipage}[t]{0.258\textwidth}
\centering
\begin{tikzpicture}
\begin{axis}[
  fast ablation compact,
  ymin=.40,
  ymax=1.02,
  ytick={.4,.6,.8,1.0},
  fast shortened y break,
]
\addplot+[fast sft bars] coordinates
  {(Chengdu,.759) (Xi'an,.784) (Porto,.436)};
\addplot+[fast grpo bars] coordinates
  {(Chengdu,.950) (Xi'an,.972) (Porto,.722)};
\end{axis}
\end{tikzpicture}
\par\vspace{-5pt}{\small\bfseries (c) Participant identification\strut}
\end{minipage}%
\hspace{0.001\textwidth}%
\begin{minipage}[t]{0.258\textwidth}
\centering
\begin{tikzpicture}
\begin{axis}[
  fast ablation compact,
  ymin=.30,
  ymax=.65,
  ytick={.3,.4,.5,.6},
  fast shortened y break,
]
\addplot+[fast sft bars] coordinates
  {(Chengdu,.470) (Xi'an,.428) (Porto,.340)};
\addplot+[fast grpo bars] coordinates
  {(Chengdu,.609) (Xi'an,.555) (Porto,.597)};
\end{axis}
\end{tikzpicture}
\par\vspace{-5pt}{\small\bfseries (d) Road-segment localization\strut}
\end{minipage}%
}
\vspace{-1pt}
\caption{Ablation of TrajMind$_{\mathrm{fast}}$. We compare the SFT initialization of TrajMind$_{\mathrm{fast}}$ with its GRPO
 refinement on (a) three-way anomaly typing (balanced accuracy),
(b) binary anomaly detection (balanced accuracy), (c) participant identification (subject F1), and
(d) road-segment localization (segment F1). GRPO improves all four metrics across three different cities.}
\Description{Four side-by-side grouped bar charts compare SFT and GRPO
across Chengdu, Xi'an, and Porto for three-way typing, binary detection,
participant identification, and road-segment localization.}
\label{fig:fast-ablation}
\label{fig:fast-detection-ablation}
\label{fig:fast-localization-ablation}
\end{figure*}

\else

\subsection{Ablation on \texorpdfstring{TrajMind$_{\mathrm{fast}}$}{TrajMind-fast} (RQ5, Figure~\ref{fig:fast-ablation})}
\label{sec:fast-ablation}

To answer \textbf{RQ5}, we compare the SFT initialization with its GRPO
refinement in \textbf{Figure~\ref{fig:fast-ablation}}. GRPO improves all four metrics across all three cities: three-way balanced accuracy rises by \(.038\)
and binary detection by \(.065\) in each city, while participant F1 gains
\(.191\), \(.188\), and \(.286\), and segment F1 gains \(.139\), \(.127\),
and \(.257\), on Chengdu, Xi'an, and Porto, respectively. The larger detection
gain suggests that reinforcement
learning particularly strengthens the fast path's primary role as a
normal--anomaly screener. For structured localization, coverage increases by
\(.191\), \(.187\), and \(.290\), closely tracking participant-F1 gains,
whereas answered-case tIoU changes only from \(.800\), \(.799\), and \(.724\)
to \(.791\), \(.787\), and \(.717\). Thus, GRPO mainly reduces missing or invalid outputs and
improves spatial localization rather than sharpening the temporal boundaries of
already answered cases. Porto obtains the largest localization gains, yet its
final participant F1 and coverage remain \(.722\) and \(.730\), below Chengdu and
Xi'an. Overall, reinforcement refinement makes
TrajMind$_{\mathrm{fast}}$ a more reliable cross-city front-line monitor,
while the remaining fine-grained typing and transfer gaps still justify the
slow path for complete diagnosis.

\fi

\FastAblationFigureOnlyfalse

%
\subsection{Analysis of Cross-Protocol Localization}
\label{sec:cross-protocol-localization}

\begin{table}[tb]
\centering
\caption{Cross-protocol anomaly localization (Subject-F1).  TrajMind is trained on Chengdu with Protocol-A. Protocols~B and~C inject the same two anomaly structures more and less strongly than the training protocol~A.}
\label{tab:cross-protocol-localization}
\small
\setlength{\tabcolsep}{5pt}
\renewcommand{\arraystretch}{0.94}
\begin{tabular}{llccc}
\toprule
City & Protocol & Derive & Cascade & TrajMind$_{\mathrm{slow}}$ \\
\midrule
Chengdu & A (training) & .640 & .761 & \textbf{.993} \\
        & B (stronger) & .719 & .805 & \textbf{.995} \\
        & C (weaker)   & .447 & .614 & \textbf{.972} \\
\addlinespace[2.5pt]
Xi'an   & A (training) & .526 & .794 & \textbf{.995} \\
        & B (stronger) & .645 & .842 & \textbf{.930} \\
        & C (weaker)   & .400 & .708 & \textbf{.914} \\
\addlinespace[2.5pt]
Porto   & A (training) & .510 & .803 & \textbf{.941} \\
        & B (stronger) & .576 & .838 & \textbf{.942} \\
        & C (weaker)   & .387 & .709 & \textbf{.941} \\
\bottomrule
\end{tabular}

\end{table}

To answer \textbf{RQ3}, we apply the same Chengdu Protocol-A localizer to two unseen
anomaly-strength protocols and city shift (on Xi'an and Porto). \textbf{Table~\ref{tab:cross-protocol-localization}}
yields three observations.

\textbf{First, TrajMind$_{\mathrm{slow}}$ consistently provides the most
accurate participant localization.}  It ranks first in all nine settings: averages \(.958\) Subject-F1, compared with \(.764\) for Cascade and \(.539\)
for Derive.  More notably, its lowest score, \(.914\) on Xi'an under the weak
Protocol~C, remains \(.072\) above the best score attained by either baseline
in any setting (Cascade's \(.842\) on Xi'an under Protocol~B).  

\textbf{Second, the symbolic methods are substantially more sensitive to
anomaly strength.}  In every city, Derive and Cascade follow the strict order
\(\mathrm{C}<\mathrm{A}<\mathrm{B}\): stronger perturbations make their fixed
thresholds easier to trigger, whereas weaker perturbations sharply reduce
localization accuracy.  On Chengdu, where only the protocol changes, moving
from Protocol~B to C lowers Subject-F1 by \(.272\) for Derive and \(.191\) for
Cascade, but by only \(.023\) for TrajMind$_{\mathrm{slow}}$.  Consequently,
the slow path's average advantage over Cascade grows from \(.127\) under the
strong Protocol~B to \(.265\) under the weak Protocol~C. This indicates that TrajMind is robust to anomaly-strength shifts, capturing the relational pattern.

\textbf{Third, robustness to protocol shift persists under simultaneous city
shift, although the interaction is city dependent.}  Relative to Protocol~A,
TrajMind$_{\mathrm{slow}}$ changes by only \(+.002/-.021\) under Chengdu's
stronger/weaker protocols.  Under the compound shift, it remains essentially
unchanged on Porto (\(.941/.942/.941\) for A/B/C), while Xi'an decreases from
\(.995\) to \(.930\) and \(.914\).  Even in the latter case, it retains a
clear lead over Cascade (\(.842\) and \(.708\)).  These results thus
support transfer across unseen anomaly strengths and cities.
\subsection{Analysis of Cross-City/Protocol Typing}
\label{sec:cross-city-protocol-typing}

\begin{table}[tb]
\centering
\caption{Anomaly typing robustness of the Chengdu-trained 3B TrajMind$_{\mathrm{slow}}$ across cities and anomaly protocols.}
\label{tab:cross-city-protocol-typing}
\small
\setlength{\tabcolsep}{5pt}
\renewcommand{\arraystretch}{0.96}
\begin{tabular}{>{\raggedright\arraybackslash}p{0.46\columnwidth}cc}
\toprule
Held-out condition & Stat.\ rule$^{a}$ & Slow (3B) \\
\midrule
\multicolumn{3}{l}{\textit{Balanced accuracy}~\(\uparrow\) } \\
\addlinespace[1pt]
\quad New city: Xi'an        & .535 & \textbf{.954} \\
\quad New city: Porto        & .533 & \textbf{.927} \\
\quad Stronger anomalies (B) & .712 & \textbf{.991} \\
\quad Weaker anomalies (C)   & .413 & \textbf{.982} \\
\quad Xi'an $+$ weaker (C)   & .409 & \textbf{.940} \\
\quad Porto $+$ weaker (C)   & .392 & \textbf{.919} \\
\addlinespace[3pt]
\multicolumn{3}{l}{\textit{False-positive rate}~\(\downarrow\)} \\
\addlinespace[1pt]
\quad Structure-only control$^{b}$ & 76.7\% & \textbf{0.0\%} \\
\bottomrule
\end{tabular}

\vspace{2pt}
\begin{minipage}{\columnwidth}
\(^{a}\)The statistical rule is training-free.  The learned column uses
Qwen2.5-VL-3B-Instruct throughout.
\(^{b}\)The structure-only control is measured on the \emph{Chengdu} test
split.
\end{minipage}
\end{table}

To answer \textbf{RQ3}, \textbf{Table~\ref{tab:cross-city-protocol-typing}}
shows that the Chengdu-trained 3B typer reaches \(.954/.927\) across cities,
\(.991/.982\) across anomaly strengths, and \(.940/.919\) under their compound
shift, versus \(.409/.392\) for the statistical rule.  Its \(0.0\%\)
false-positive rate on structure-only controls, compared with \(76.7\%\) for
the rule, confirms reliance on temporal evidence rather than co-routing.

\begin{figure*}[t]
\centering
\begin{minipage}[c]{0.238\textwidth}
  \centering
  \includegraphics[height=1.24in,trim=10bp 5bp 8bp 5bp,clip]{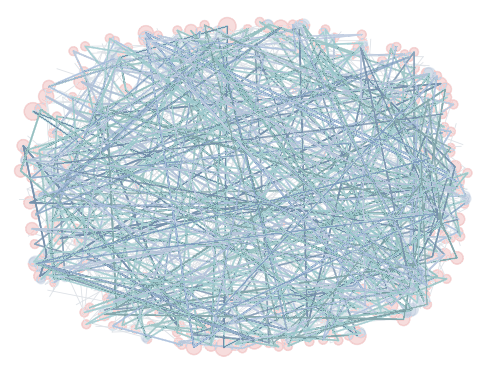}
\end{minipage}%
\hspace{0.0035\textwidth}%
\begin{minipage}[c]{0.238\textwidth}
  \centering
  \includegraphics[height=1.24in,trim=12bp 6bp 10bp 6bp,clip]{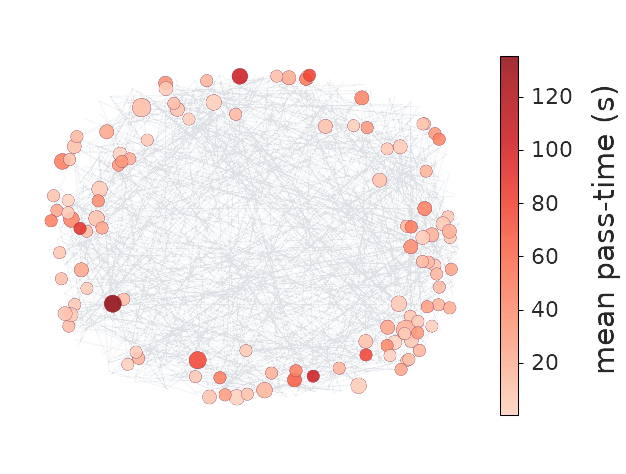}
\end{minipage}%
\hspace{0.022\textwidth}%
\begin{minipage}[c]{0.238\textwidth}
  \centering
  \includegraphics[height=1.24in,trim=10bp 5bp 8bp 5bp,clip]{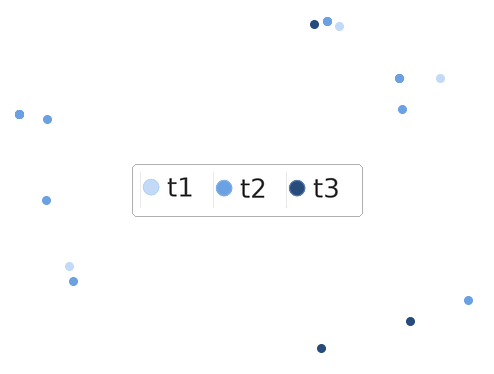}
\end{minipage}%
\hspace{0.022\textwidth}%
\begin{minipage}[c]{0.238\textwidth}
  \centering
  \includegraphics[height=1.24in,trim=2.5bp 1.2bp 2bp 1.2bp,clip]{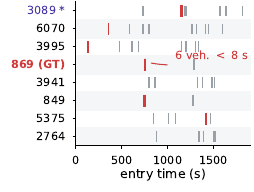}
\end{minipage}
\caption{A Bunching Anomaly Instance. From left to right: (a)
trajectories/co-occurrence, (b) mean pass time, (c) temporal group formation,
and (d) busy-segment arrivals. In~(d), the six injected vehicles (red) enter
ground-truth segment~869 within 8\,s; ``*'' marks prediction~3089.}
\Description{Four panels in a row. The first shows dense per-vehicle trajectories
over a road network, the second a mean pass-time map, the third a temporal
storyboard of group formation, and the fourth a strip of vehicle entry times
on eight shared road segments, with six red ticks clustered within eight
seconds on the ground-truth segment 869.}
\label{fig:case-study-compact}
\end{figure*}


\begin{table}[tb]
\centering
\caption{Inference latency of TrajMind with Qwen2.5-VL-3B-Instruct on one NVIDIA H20, using 179 Chengdu windows.
}  
\label{tab:inference-time}
\small
\setlength{\tabcolsep}{7pt}
\renewcommand{\arraystretch}{0.92}
\begin{tabular}{lr}
\toprule
Method & Latency (s/window) \\
\midrule
\multicolumn{2}{l}{\textit{End-to-end deployment }} \\
\quad \(\mathrm{TrajMind}_{\mathrm{fast}}\) & 5.315 \\
\quad \(\mathrm{TrajMind}_{\mathrm{slow}}\) & 9.026 \\
\quad Slow--fast difference & 3.711 \\
\quad Slow/fast ratio & \(1.70\times\) \\
\bottomrule
\end{tabular}

\end{table}

\subsection{Inference Time Comparison (RQ4, Table~\ref{tab:inference-time})}
\label{sec:inference-time}

To answer \textbf{RQ4}, \textbf{Table~\ref{tab:inference-time}} compares the two deployed
pipelines under the same batch-one generation budget.  TrajMind$_{\mathrm{fast}}$
requires \(5.315\)\,s per window, versus \(9.026\)\,s for
TrajMind$_{\mathrm{slow}}$, saving \(3.711\)\,s (\(41.1\%\)) by avoiding
canvas construction and a separate localization pass.  This saving largely
preserves in-domain screening: on Chengdu, the fast and slow paths obtain
\(.960\) versus \(.974\) binary balanced accuracy and \(.950\) versus \(.993\) subject
F1, respectively, although the fast path is notably weaker at fine-grained
typing (\(.637\) versus \(.960\)).  The trade-off widens under domain shift:
on Porto, its subject F1 and coverage fall to \(.722\) and \(.730\), compared
with \(.941\) and \(1.000\) for the slow path
(Tables~\ref{tab:t7-localization} and~\ref{tab:cross-city-localization}).
Therefore, the fast path is suited to latency-sensitive screening, whereas the
slow path remains preferable for complete and fine-grained diagnosis.

\subsection{Ablation on Specialized Roles (RQ5, Table~\ref{tab:role-specialization})}
\label{sec:role-specialization}

\begin{table}[t]
\centering
\caption{Role-specialization ablation. Results follow CD/XA/PT.
\textbf{Bold} marks the best result and SoLA denotes the shared LoRA baseline.}
\label{tab:role-specialization}
\small
\setlength{\tabcolsep}{4.5pt}
\renewcommand{\arraystretch}{0.92}
\begin{tabular}{lccc}
\toprule
Variant & \shortstack{Three-way typing} & Subject F1 & Segment F1 \\
\midrule
SoLA$_{\mathrm{3B},r=16}$
  & .855/.836/.792 & .925/.916/.852 & .402/.486/.793 \\
SoLA$_{\mathrm{3B},r=32}$
  & .873/.848/.843 & .932/.935/.882 & .415/.504/.825 \\
TrajMind$_{\mathrm{3B, slow}}$
  & \textbf{.960/.954/.927}
  & \textbf{.993/.995/.941}
  & \textbf{.502/.636/.930} \\
\bottomrule
\end{tabular}
\end{table}

The two Shared LoRA baselines use one 3B backbone and one LoRA shared by canvas-typing
and text-localization; $r{=}32$ matches the aggregate adapter budget of the two
$r{=}16$ roles it replaces (the typer and the localizer).  Raising the shared rank from 16
to 32 improves typing by \(.012\)--\(.051\), subject F1 by \(.007\)--\(.030\),
and segment F1 by \(.013\)--\(.032\), with the largest gains on Porto.  Thus,
capacity helps but only partially closes the gap.  Under the parameter-matched
comparison, TrajMind$_{\mathrm{slow}}$ still gains \(.087/.106/.084\) in typing,
\(.061/.060/.059\) in subject F1, and \(.087/.132/.105\) in segment F1 on
CD/XA/PT.  These consistent cross-task and cross-city gains support reduced
interference between global canvas recognition and exact lossless-text recovery,
rather than merely allocating more rank.  The largest parameter-matched
segment gain occurs on Xi'an (\(.132\)), indicating that shared adaptation is
especially limiting for exact spatial grounding under cross-city shift.  

\begin{table*}[t]
\centering
\begin{minipage}[t]{0.485\textwidth}
\centering
\captionof{table}{Zero-shot canvas typing on Chengdu. Two-way excludes
\emph{normal}; three-way is the deployment task. Norm./slow./bunch. report
per-class recall.}
\label{tab:zeroshot-calibration}
\footnotesize
\setlength{\tabcolsep}{4pt}
\renewcommand{\arraystretch}{0.88}
\begin{tabular}{lrrrrr}
\toprule
Model & 3-way & 2-way & norm. & slow. & bunch. \\
\midrule
Qwen3-VL-2B & .333 & .531 & .00 & .00 & 1.00 \\
Qwen2.5-VL-3B & .333 & .520 & .00 & .00 & 1.00 \\
Qwen2.5-VL-7B & .522 & \textbf{.983} & .99 & .25 & .33 \\
Qwen3-VL-8B & .670 & \textbf{.994} & .01 & 1.00 & 1.00 \\
\addlinespace[2pt]
Qwen3-VL-235b-a22b & .667 & \textbf{1.000} & .00 & 1.00 & 1.00 \\
Qwen3-VL-plus & .540 & .851 & .00 & .72 & 1.00 \\
GLM-5V-Turbo & .574 & \textbf{1.000} & .03 & .72 & 1.00 \\
\midrule
TrajMind$_{\mathrm{slow}}$ (3B) & \textbf{.960} & .972 & .96 & .95 & .97 \\
\bottomrule
\end{tabular}
\end{minipage}
\hfill
\begin{minipage}[t]{0.485\textwidth}
\centering
\captionof{table}{Verifier ablation on Chengdu. \emph{w/o verifier} releases
every end-to-end claim, whereas TrajMind$_{\mathrm{slow}}$ releases only
diagnoses that pass verification. $\Delta$ is relative to \emph{w/o verifier};
\textbf{bold} marks the better performance among the two.}
\label{tab:verifier-ablation}
\small
\setlength{\tabcolsep}{4.5pt}
\renewcommand{\arraystretch}{1.02}
\begin{tabular}{lccr}
\toprule
Metric & w/o verifier & TrajMind$_{\mathrm{slow}}$ & $\Delta$ (\%) \\
\midrule
Three-way typing accuracy & .960          & \textbf{.990} & $+3.1$  \\
Participant F1            & .960          & \textbf{.996} & $+3.8$  \\
Segment F1                & .480          & \textbf{.530} & $+10.4$ \\
tIoU (answered only)      & .750          & \textbf{.760} & $+1.3$  \\
Released coverage         & \textbf{.960} & .930          & $-3.1$  \\
\bottomrule
\end{tabular}
\vskip -0.2in
\end{minipage}
\end{table*}


%
%
\subsection{Analysis of Zero-Shot Failure: Perception or Calibration? (RQ6, Table~\ref{tab:zeroshot-calibration})}
\label{sec:zeroshot-calibration}

Zero-shot canvas reading tops out well below the adapted model.  Because the deployment
argument rests on training a small adapter rather than prompting a larger model, it matters
whether the zero-shot ceiling is a perceptual limit or a decision-boundary one. 
\textit{The zero-shot ceiling is primarily a calibration failure.}
Removing \emph{normal} raises the capable frozen models to near-ceiling two-way
accuracy (\(.983\)--\(1.000\)), whereas their three-way balanced accuracy remains
between \(.522\) and \(.670\).  Their class recalls reveal opposing biases: the 7B
model nearly always predicts \emph{normal} (\(.99\) recall) and misses most
slowdowns, while the 8B and 235B models recognize both anomaly types but almost
never predict \emph{normal}.  Thus, the collective patterns are perceptually
available, but the normal--anomaly boundary is poorly calibrated; scaling from 8B
to 235B does not resolve it.  In contrast, TrajMind$_{\mathrm{slow}}$ reaches
\(.960\) three-way balanced accuracy with balanced per-class recalls
(\(.96/.95/.97\)).  The opposing recall profiles further show that the
zero-shot errors are systematic operating-point shifts rather than uniformly
uncertain predictions.  Consequently, aggregate accuracy alone can overstate
diagnostic readiness unless normal-class recall is examined together with
anomaly-subtype recall. 
The result supports
adapting a compact canvas-typing role for learnable calibration, rather than
claiming unrestricted generalization or merely scaling zero-shot VLMs.

\subsection{Verifier Ablation (Table~\ref{tab:verifier-ablation})}
\label{sec:verifier-ablation}

To isolate the contribution of verification, we compare the Chengdu slow path
with the verifier disabled and enabled.  Metrics for the verified configuration
are computed over released diagnoses and should therefore be interpreted
together with released coverage.

Table~\ref{tab:verifier-ablation} shows a consistent improvement in the
quality of released diagnoses.  Verification raises three-way typing accuracy
from $.960$ to $.990$ and participant F1 from $.960$ to $.996$.  The largest
relative gain appears in segment localization, where F1 increases from $.480$
to $.530$ ($+10.4\%$), whereas answered-only tIoU improves more modestly from
$.750$ to $.760$ ($+1.3\%$).  These gains reduce released coverage from $.960$
to $.930$, a three-percentage-point decrease.  Thus, the verifier improves
reliability by withholding a small additional subset of diagnoses, with a
larger benefit for participant and segment selection than for temporal-boundary
precision.  Viewed as an operating-point trade-off, the \(.030\) coverage loss
is small relative to the \(.050\) absolute gain in segment F1, suggesting that
rejected cases are disproportionately spatially unreliable.  Accordingly, the
verification threshold can serve as a deployment control for balancing answer
availability against the cost of releasing an unsupported localization.
Because it verifies or withholds model outputs rather than
rewriting them, these improvements should be interpreted as selective release
rather than error correction.

\subsection{Case Study: A Bunching Anomaly Instance}
\label{sec:case-study-compact}

\textbf{Figure~\ref{fig:case-study-compact}} traces a {900\,s} Chengdu window
containing 54 trajectories.  Panel~(a) shows the dense route interactions,
panel~(b) summarizes mean pass time on the shared road segments, and panel~(c)
reveals how the candidate group forms over time.  Panel~(d) then exposes the
decisive temporal signature: all six injected vehicles enter ground-truth
segment~869 within 8.5\,s.  This compact arrival cluster supports bunching
rather than a broad slowdown, while the remaining vehicles are dispersed over
the window.

The stage outputs separate \emph{what}, \emph{who}, and \emph{where}.  The fast
path recovers all six injected identities but labels the event as a slowdown
and points to segment~3089.  The slow canvas corrects the type to
\texttt{anomaly\_bunching} with confidence~1.0 and retains the same six
participants, yet its text localizer still selects segment~3089.  Recomputing
that claim against the lossless trajectories shows that only three of the six
vehicles traverse 3089, whereas all six pass through segment~869.  The
\texttt{passed\_through} check therefore returns contradiction~\(.661\), and
the verifier rejects the diagnosis despite strong group-level consistency.
It does not repair the segment or replace the hypothesis. Instead, it withholds an
unsupported localization.  The failed check also provides an explicit reason
for abstention.  The case consequently shows why correct event
typing and participant recovery do not guarantee correct spatial grounding,
which is also the main source of the remaining segment-F1 gap. More broadly, this disagreement localizes the remaining failure: the
representation captures collective membership and temporal synchronization, but segment selection remains brittle.  
Verification prevents a correct high-level diagnosis from legitimizing an unsupported claim.


\section{Conclusion}

In this paper, we propose TrajMind, a fast-and-slow framework for collective
trajectory anomaly detection and diagnosis. TrajMind specializes a shared
frozen vision--language backbone with role-specific LoRA adapters. Its slow
path combines canvas-based typing, lossless textual localization, and evidence
verification, while its fast path provides single-pass text-only screening.
Experiments on Chengdu, Xi'an, and Porto show that the slow path delivers
accurate, evidence-backed diagnoses across cities and anomaly-severity levels.
The fast path reduces latency while preserving detection and participant
identification.



\section{Ethical Considerations}
\label{sec:ethics}

Trajectory analysis can expose sensitive mobility patterns and can be
misused for pervasive surveillance or automated enforcement.  Although the
present study addresses group-level traffic behavior rather than demographic
profiling, participant identifiers and fine-grained spatiotemporal records
may still enable re-identification when combined with external data.  A
deployment should therefore apply data minimization, pseudonymization,
retention limits, access control, and auditing consistent with the governing
privacy and transportation regulations.

Role-specific fine-tuning and compact training data introduce distinct model
risks. Even benign fine-tuning data can erode an aligned model's safeguards,
motivating safety-aware control of parameter updates~\citep{lu2025safedelta};
data-condensation pipelines can also preserve injected backdoors while
retaining apparent utility~\citep{wu2025backdoor}. Moreover, goal and resource
pressure can cause language agents to trade safety constraints for task
completion~\citep{jiang2026agentspressure}. Accordingly, neither the fast
path's throughput objective nor the slow path's diagnosis objective should
override fixed deployment constraints. Adapter provenance, training-data
integrity, and safety behavior should be audited before deployment and after
each update.


\bibliographystyle{ACM-Reference-Format}
\bibliography{references}

\end{document}